\documentclass{article}
\usepackage{arxiv}

\usepackage[utf8]{inputenc} 
\usepackage[T1]{fontenc}    
\usepackage{url}            
\usepackage{booktabs}       
\usepackage{amsfonts}       
\usepackage{nicefrac}       
\usepackage{microtype}      
\usepackage{lipsum}
\usepackage{cellspace}
\usepackage{alltt}
\usepackage{calc}
\usepackage{mathtools}
\usepackage{booktabs}
\usepackage{url}
\usepackage{caption}
\usepackage{longtable}
\usepackage{amsmath}
\usepackage{amsthm}
\usepackage{amssymb}
\usepackage{bm}
\usepackage{adjustbox}
\usepackage{wrapfig}
\usepackage{epstopdf}
\usepackage{tikz}
\usepackage{framed}
\usepackage{thmtools,thm-restate}
\usepackage{enumitem}
\usepackage[ruled]{algorithm2e}
\usepackage{soul}
\usepackage{tcolorbox}
\usepackage{multirow}
\usepackage{array}
\usepackage{makecell}

\usepackage{lipsum}

\usepackage[colorlinks=true,breaklinks]{hyperref}
\usepackage{cleveref}

\usetikzlibrary{arrows}
\usetikzlibrary{shapes,backgrounds}

\graphicspath{{./Images/}}

\newtheorem*{theorem*}{Theorem}

\newtheorem*{lemma*}{Lemma}

\makeatletter
\renewcommand{\th@definition}{%
  \normalfont
  \thm@preskip-2 \relax
  \thm@postskip-2 \relax
}
\makeatother

\title{Regularized Optimal Transport for Dynamic  Semi-supervised Learning}

\author{Mourad El Hamri, Youn{\`e}s Bennani\\
  Universit\'e Sorbonne Paris Nord, LIPN 7030 UMR CNRS\\
  LaMSN - La Maison des Sciences Numériques\\
  \url{name.surname@sorbonne-paris-nord.fr}}

\begin{document}
\maketitle

\newcommand\nnfootnote[1]{%
  \begin{NoHyper}
  \renewcommand\thefootnote{}\footnote{#1}%
  \addtocounter{footnote}{-1}%
  \end{NoHyper}
}
\nnfootnote{Paper submitted to Pattern Recognition}

\begin{abstract}
Semi-supervised learning provides an effective paradigm for leveraging unlabeled data to improve a model’s performance. Among the many strategies proposed, graph-based methods have shown excellent properties, in particular since they allow to solve directly the transductive tasks according to Vapnik's principle and they can be extended efficiently for inductive tasks. In this paper, we propose a novel approach for the transductive semi-supervised learning, using a complete bipartite edge-weighted graph. The proposed approach uses the regularized optimal transport between empirical measures defined on labelled and unlabelled data points in order to obtain an affinity matrix from the optimal transport plan. This matrix is further used to propagate labels through the vertices of the graph in an incremental process ensuring the certainty of the predictions by incorporating a certainty score based on Shannon’s entropy. We also analyze the convergence of our approach and we derive an efficient way to extend it for out-of-sample data. Experimental analysis was used to compare the proposed approach with other label propagation algorithms on 12 benchmark datasets, for which we surpass state-of-the-art results. We release our code. \footnote{Code is available at \url{https://github.com/MouradElHamri/OTP}} 
\end{abstract}



\section{Introduction}
\label{intro}
Semi-supervised learning has recently emerged as one of the most promising paradigms to mitigate the reliance of deep learning on huge amounts of labeled data, especially in learning tasks where it is costly to collect annotated data. This is best illustrated in medicine, where measurement require overpriced machinery and labels are the result of an expensive human assisted time-consuming analysis. 
\newline
\newline 
Semi-supervised learning (SSL) aims to largely reduce the need for massive labeled datasets by allowing a model to leverage both labeled and unlabeled data. Among the many semi-supervised learning approaches, graph-based semi-supervised learning techniques are increasingly being studied due to their performance and to more and more real graph datasets. The problem is to predict all the unlabelled vertices in the graph based on only a small subset of vertices being observed. To date, a number of graph-based algorithms, in particular label propagation methods have been successfully applied to different fields, such as social network analysis  \cite{boldi2011layered}\cite{xie2013labelrank}\cite{zhang2017label}\cite{jokar2019community}, natural language processing \cite{alexandrescu2007data}\cite{tamura2012bilingual}\cite{barba2020mulan}, and image segmentation \cite{wang2007efficient}\cite{breve2019interactive}.
\newline
\newline The performance of label propagation algorithms is often affected by the graph-construction method and the technique of inferring pseudo-labels. However, in the graph-construction, traditional label propagation approaches are incapable of exploiting the underlying geometry of the whole input space, and the entire relations between labelled and unlabelled data in a global vision. Indeed, authors in \cite{zhu2002learning}\cite{zhou2003learning} have adopted pairwise relationships between instances by relying on a Gaussian function with a free parameter $\sigma$, whose optimal value can be hard to determine in certain scenarios, and even a small perturbation in its value can affect significantly the classification results. Authors in \cite{wang2007label} have suggested to derive another way to avoid the use of $\sigma$, by relying on the local concept of linear neighborhood, though, the linearity hypothesis is intended just for computational convenience, a variance in the size of the neighborhood can also change drastically the classification results. Moreover, these algorithms have the inconvenience of inferring pseudo-labels by hard assignment, ignoring the different degree of certainty of each prediction. The performance of these label propagation algorithms can also be judged by its generalization ability for out-of-sample data. An efficient label propagation algorithm capable of addressing all these points has not yet been reported.
\newline
\newline One of the paradigms used to capture the underlying geometry of the data is grounded on the theory of optimal transport \cite{villani2008optimal}\cite{santambrogio2015optimal}. Optimal  transport provides a means of lifting distances between data in the input space to distances between probability distributions over this space. The latter are called the Wasserstein distances, and they have many favorable theoretical properties, that have attracted a lot of attention for machine learning tasks such as domain adaptation \cite{courty2014domain}\cite{courty2016optimal}\cite{NIPS2017_0070d23b}\cite{bhushan2018deepjdot}\cite{redko2019optimal}, clustering \cite{laclau2017co}\cite{chakraborty2020hierarchical}\cite{bouazza2019multi}\cite{bouazza2020collaborative}, generative models \cite{martin2017wasserstein}\cite{salimans2018improving} and image processing \cite{rabin2011wasserstein}\cite{de2012blue}.
\newline
\newline Recently, optimal transport theory has found renewed interest in semi-supervised learning. In \cite{solomon2014wasserstein}, the authors provided an efficient approach to graph-based semi-supervised learning of probability distributions. In a more recent work, the authors in \cite{taherkhani2020transporting} adopted a hierarchical optimal transport technique to find a mapping from empirical unlabeled measures to corresponding labeled measures. Based on this mapping, pseudo-labels for the unlabeled data are inferred, which can then be used in conjunction with the initial labeled data to train a CNN model in an SSL manner.
\newline
\newline Our solution to the problems above, presented in this paper, is to propose a novel method of semi-supervised learning, based on optimal transport theory, called OTP : Optimal Transport Propagation. OTP presents several differentiating points in relation to the state of the art approaches. We can summarize its main contributions as follows:
\begin{itemize}
    \item Construct an enhanced version of the affinity matrix to benefit from all the geometrical information in the input space and the interactions between labelled and unlabelled data using the solid mathematical framework of optimal transport.
    \item The use of uncommon type of graph for semi-supervised learning : a complete bipartite edge-weighted graph, to avoid working within the hypothesis of label noise, thus there is no more need to add an additional term in the objective function in order to penalize predicted labels that do not match the correct ones.
    \item The use of an incremental process to take advantage of the dependency of semi-supervised algorithms on the amount of prior information, by increasing the number of labelled data and decreasing the number of unlabelled ones at each iteration during the process.
    \item Incorporate a certainty score based on Shannon's entropy to control the certitude of our predictions during the incremental process.
    \item OTP can be efficiently extended to out-of-sample data in multi-class inductive settings : Optimal Transport Induction (OTI).
\end{itemize}
The  remainder  of  this  paper  is  organized  as  follows: we provide  background  on optimal transport in Section 2, followed by an overview of semi-supervised learning in Section 3. Novel contributions are presented in Section 4. We conclude with experimental results on real world data in Section 5, followed by the conclusion and future works in Section 6.
\section{Optimal transport}
\label{sec:1}
\textbf{Monge's problem :} The birth of optimal transport theory dates back to 1781, when the French mathematician Gaspard Monge \cite{monge1781memoire} introduced the following problem : Given two probability measures $\mu$ and $\nu$ over metric spaces $\mathcal{X}$ and $\mathcal{Y}$ respectively, and a measurable cost function $c : \mathcal{X}\times\mathcal{Y} \to \mathbb{R}^+ $, which represents the work needed to move one unit of mass from location $x \in \mathcal{X}$ to location  $y \in \mathcal{Y}$, the problem asks to find a transport map $\mathcal{T} : \mathcal{X} \to \mathcal{Y}$, that transforms the mass represented by the measure $\mu$, to the mass represented by the measure $\nu$, while minimizing the total cost of transportation, i.e. Monge's problem yields a map $\mathcal{T}$ that realizes :
\begin{equation} (\mathcal{M})  \quad \underset{\mathcal{T}}{\inf}\{\int_{\mathcal{X}}  c(x,\mathcal{T}(x)) d\mu(x) |  \mathcal{T}\#\mu = \nu \},   \end{equation}
where $\mathcal{T}\#\mu$ denotes the measure image of $\mu$ by $\mathcal{T}$, defined by : for all measurable subset $\mathcal{B} \subset \mathcal{Y}$, $\mathcal{T}\#\mu(\mathcal{B}) = \mu(\mathcal{T}^{-1}(\mathcal{B})) = \mu(\{x \in \mathcal{X} : \mathcal{T}(x) \in \mathcal{B}\})$. The condition $\mathcal{T}\#\mu = \nu$, models a local mass conservation constraint : the amount of masses received in any subset $\mathcal{B} \subset \mathcal{Y}$ corresponds to what was transported here, that is the amount of masses initially present in the pre-image $\mathcal{T}^{-1}(\mathcal{B})$ of $\mathcal{B}$ under $\mathcal{T}$.
\newline
\begin{figure}[h]
	\centering
\includegraphics[width=0.9\textwidth]{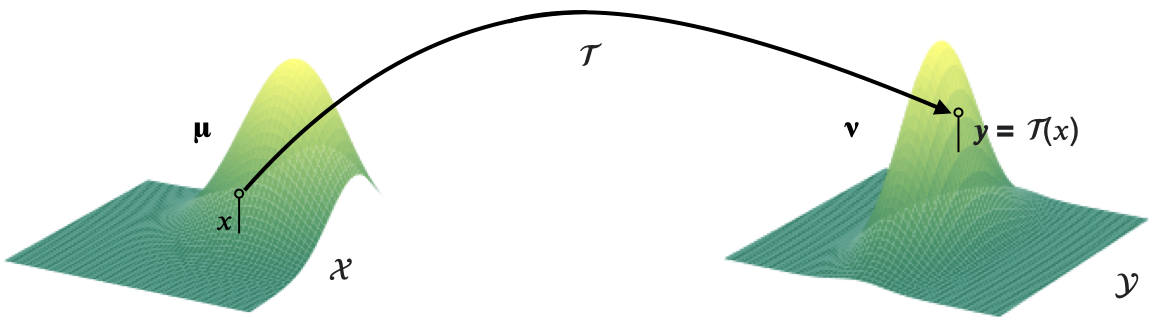}
	\label{fig:pcs}
	\caption{Monge’s problem. $\mathcal{T}$ is a transport map from $\mathcal{X}$ to $\mathcal{Y}$ }
\end{figure}
\newline The solution to $(\mathcal{M})$ might not exist because of the very restrictive local mass conservation constraint, it is the case, for instance when $\mu$ is a Dirac mass and $\nu$ is not. One also remarks that $(\mathcal{M})$ is rigid in the sense that all the mass at $x$ must be associated to the same target $y=\mathcal{T}(x)$. In fact, it is clear that if the mass splitting really occurs, which means that there are multiple destinations for the mass at $x$, then this displacement cannot be described by a map $\mathcal{T}$. Moreover, the problem $(\mathcal{M})$ is not symmetrical, the measures $\mu$ and $\nu$ do not play the same role, and this causes additional difficulties when studying the existence of solutions for problem $(\mathcal{M})$.
\newline
\newline \textbf{Kantorovich’s relaxed problem :} The problem of Monge has stayed with no solution until 1942, when the Soviet mathematician and economist Leonid Kantorovitch \cite{kantorovich1942translocation} suggested a convex relaxation of $(\mathcal{M})$, which allows mass splitting and guaranteed to have a solution under very general  assumptions, this relaxed formulation is known as the Monge-Kantorovich problem :
\begin{equation} (\mathcal{MK}) \,\,\,\,\,\,\underset{\gamma}{\inf} \{\, \int_{\mathcal{X}\times\mathcal{Y}} \, c(x,y) \, d\gamma(x,y) \,|\, \gamma \in \Pi(\mu,\nu)\, \}, \end{equation}
where $\Pi(\mu,\nu)$ is the set of transport plans constituted of all joint probability measures $\gamma$ on $\mathcal{X}\times\mathcal{Y}$ with marginals $\mu$ and $\nu$ :
\begin{center}
  $\Pi(\mu,\nu) = \{ \gamma \in \mathcal{P}(\mathcal{X}\times\mathcal{Y})|\pi_1\#\gamma = \mu$ and $\pi_2\#\gamma = \nu$\}     
\end{center}
${\pi}_{1}$ and ${\pi}_{2}$ stand for the projection maps : $\begin{aligned}[t]
\pi_1 \colon \mathcal{X}\times\mathcal{Y} &\to \mathcal{X} \\
(x,y) &\mapsto x 
\end{aligned}
\quad\text{and}\quad
\begin{aligned}[t]
\pi_2 \colon \mathcal{X}\times\mathcal{Y} &\to \mathcal{Y}. \\
(x,y) &\mapsto y
\end{aligned}$
\newline
\newline The main idea of Kantorovich is to widen the very restrictive problem $(\mathcal{M})$, instead of minimizing the total cost of the transport according to the map $\mathcal{T}$, it is towards probability measures over the product space $\mathcal{X}\times\mathcal{Y}$ that Kantorovitch's look turns to, these joint probability measures are a different way to describing the displacement of the mass of $\mu$ : rather than specifying for each $x \in \mathcal{X}$, the destination $y = \mathcal{T}(x) \in \mathcal{Y}$ of the mass originally presented at $x$, we specify for each pair $(x, y) \in  \mathcal{X}\times\mathcal{Y}$ how much mass goes from $x$ to $y$. Intuitively, the mass initially present at $x$ must correspond to the sum of the masses “leaving” from $x$ during the transportation, and in a similar manner, the final mass prescribed in $y$ should correspond to the sum of the masses “arriving” in $y$, that can be written as : $\mu(x)= \int_{\mathcal{Y}}\,d\gamma(x,y)$ and $\nu(y)= \int_{\mathcal{X}}\,d\gamma(x,y)$, which corresponds to a condition on the marginals : $\pi_1\#\gamma = \mu$,  and $\pi_2\#\gamma = \nu$, we are therefore now limited to working on measures whose marginals coincide with $\mu$ and $\nu$ : $\Pi(\mu,\nu)$ is then the admissible set of $(\mathcal{MK})$. The relaxation of Kantorovitch is a much more suitable framework which gives the possibility of mass splitting. Furthermore this relaxation has the virtue of guaranteeing the existence of a solution under very general assumptions : $\mathcal{X}$ and $\mathcal{Y}$ are Polish spaces, and the cost function $c$ is lower semi-continuous.
\begin{figure}[h]
\centering
\includegraphics[width=0.8\textwidth]{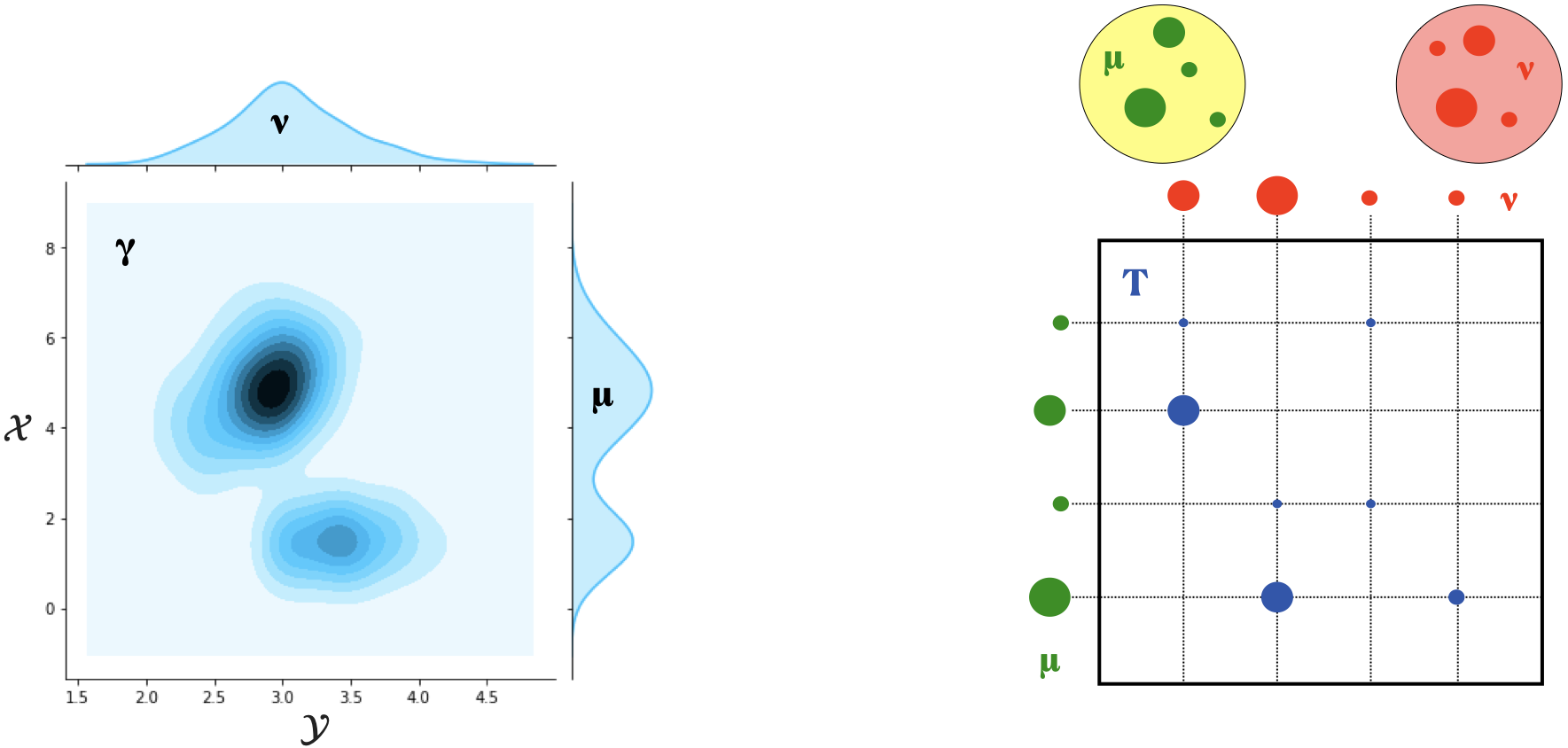}
	\label{fig:pccs}
	\caption{Continuous Kantorovich’s relaxation : the joint probability distribution $\gamma$ is a transport plan between $\mu$ and $\nu$ (left). Discrete Kantorovich's relaxation : the positive entries of the discrete transport plan are displayed as blue disks with radius proportional to the entries values (right)}
\end{figure}
\newline 
\newline \textbf{The Wasserstein distances :} Whenever $\mathcal{X}=\mathcal{Y}$ is equipped with a metric $d$, it is natural to use it as a cost function, e.g. $c(x, y) = d(x, y)^p$ for $p \in {\left[1\,,+\infty\right[}$. In such case, the problem of Monge-Kantorovich defines a distance between probability measures over $\mathcal{X}$, called the $p$-Wasserstein distance, defined as follows : 
\begin{equation}
    \mathrm{W}_{p}(\mu,\nu) = \underset{\gamma \in \Pi(\mu,\nu)}{\inf} (\int_{\mathcal{X}^{2}} \, d^{p}(x,y) \, d\gamma(x,y))^{1/p}, \,\,\,\,\, \forall \mu,\nu \in \mathcal{P}(\mathcal{X})
\end{equation}
The Wasserstein distance is a powerful tool to make meaningful comparisons between probability measures, even when the supports of the measures do not overlap. This metric has several other advantages, first of all, they have an intuitive formulation and they have the faculty to capture the underlying geometry of the measures by relying on the cost function $c = d^p$ which encodes the geometry of the space $\mathcal{X}$. The fact that the Wasserstein distance metrize weak convergence is an other major advantage and makes it an ideal candidate for learning problems.
\newline
\newline \textbf{Discrete optimal transport :} In applications, direct access to the measure $\mu$ and $\nu$ is often not available, instead, we have only access to i.i.d. finite samples from $\mu$ and $\nu$ : $X = (x_1,...,x_n) \subset \mathcal{X}$ and $Y = (y_1,...,y_m) \subset \mathcal{X}$, in that case, these can be taken to be discrete measure : $\mu = \sum_{i=1}^n a_{i} \delta_{x_{i}}$ and
$\nu = \sum_{j=1}^m b_{j} \delta_{y_{j}}$,  where $a$ and $b$ are vectors in the probability simplex : $a=(a_1,...,a_n) \in \sum_n$ and $b=(b_1,...,b_m) \in \sum_m$, where :
$\sum_k = \{ u \in \mathbb{R}^k \, | \, \forall i \le k, \, u_i \ge 0 \, \,and \, \sum_{i=1}^k   u_i =1 \}$. 
The pairwise costs can be compactly represented as an $n \times m$ matrix $M_{XY}$ of pairwise distances between elements of $X$ and $Y$ raised to the power $p$ :
\begin{center}
$M_{XY} = [d(x_i,y_j)^p]_{i,j} \in  \mathcal{M}_{n \times m}(\mathbb{R}^{+})$ 
\end{center}
In this case, the $p$-Wasserstein distance becomes the $p^{th}$ root of the optimum of a network flow problem known as the transportation problem \cite{bertsimas1997introduction}, this is a parametric linear program on $n \times m$ variables, parameterized by two elements : the matrix $M_{XY}$, which acts as a cost parameter, and the transportation polytope $U(a,b)$ which acts as a feasible set, defined as the set of $n \times m$ non-negative matrices such that their row and column marginals are equal to $a$ and $b$ respectively :
\begin{center}
   $U(a,b) = \{T \in \mathcal{M}_{n \times m}(\mathbb{R}^{+}) \, | \, T 1_{m} = a \,\, \text{and} \,\, T^{\mathbf{T}} 1_{n} = b\}$. 
\end{center}
Let $ \langle A,B \rangle _F \,= trace(A^\mathrm{T}B)$ be the Frobenius dot-product of matrices, then ${\mathrm{W}^{p}_{p}}(\mu,\nu)$ —the distance ${\mathrm{W}_{p}}(\mu,\nu)$ raised to the power $p$— can be written as :
\begin{equation}
{\mathrm{W}^{p}_{p}}(\mu,\nu) = \underset{T \in U(a,b)}{\min}  \langle {T},{M_{XY}} \rangle _F 
\end{equation}
Note that if $n = m$, and, $\mu$ and $\nu$ are uniform measures, $U(a,b)$ is then the Birkhoff polytope of size $n$, and the solutions of ${\mathrm{W}^{p}_{p}}(\mu,\nu)$, which lie in the corners of this polytope, are permutation matrices.
\newline 
\newline Discrete optimal  transport  defines a powerful framework to compare  probability measures in a geometrically faithful way. However, the impact of optimal transport in machine learning community has long been limited and neglected in favor of simpler $\varphi$-divergences or MMD because of its computational and statistical burden. in fact, optimal transport is too costly and suffers from the curse of dimensionality:
\newline Optimal transport have a heavy computational price tag : discrete optimal transport is a linear program, and thus can be solved with interior point methods using network flow solvers, but this problem scales cubically on the sample size, in fact the computational complexity is $\mathcal{O}(n^3 log(n))$ when comparing two discrete measures of $n$ points in a general metric space \cite{pele2009fast}, which is often prohibitive in practice.
\newline Optimal transport suffers from the curse of dimensionality : considering a probability measure $\mu$ over $\mathbb{R}^d$, and its empirical estimation $\hat{\mu}_n$, we have $\mathbb{E} [\mathrm{W}_{p}(\mu, \hat{\mu}_n)] = \mathcal{O}(n^{\frac{-1}{d}})$ \cite{weed2019sharp}. The empirical distribution $\hat{\mu}_n$ becomes less and less representative as the dimension $d$ of the ambient space $\mathbb{R}^d$ becomes large,  so that in the convergence of $\hat{\mu}_n$ to $\hat{\mu}$ in Wasserstein distance is slow.
\newline
\newline \textbf{Entropy-regularized optimal transport :} Entropy-regularized optimal transport has recently emerged as a solution to the computational issue of optimal transport \cite{cuturi2013sinkhorn}, and to the sample complexity properties of its solutions \cite{genevay2019sample}. 
\newline
\newline The entropy-regularized problem reads: 
\begin{equation}
\underset{\gamma \in \Pi(\mu,\nu)}{\inf} \int_{\mathcal{X}\times\mathcal{Y}} \, c(x,y) \, d\gamma(x,y) +\varepsilon \mathcal{H}(\gamma),
\end{equation}
where $\mathcal{H}(\gamma) = \int_{\mathcal{X}\times\mathcal{Y}} \, \log(\frac{d\gamma(x,y)}{dxdy}) \, d\gamma(x,y)$ is the entropy of the transport plan $\gamma$.
\newline
\newline The entropy-regularized version of the discrete optimal transport reads:
\begin{equation}
    \underset{T \in U(a,b)}{\min}  \langle {T},{M_{XY}} \rangle _F - \varepsilon \mathcal{H}(T)   ,
\end{equation}
where  $\mathcal{H}(T) = - \sum_{i=1}^n \sum_{j=1}^m t_{ij} (\log(t_{ij}) - 1)  $ is the entropy of $T$
\newline
\newline The main idea is to use $\mathcal{-H}$ as a regularization function to obtain approximate solutions to the original transport problem. The intuition behind this form of regularization is : since the cardinality of the nonzero elements of the optimal transport plan $T^*$ is at most $m + n - 1$ \cite{brualdi2006combinatorial}, one can look for a smoother version of the transport, thus lowering its sparsity, by increasing its entropy. 
\newline
\newline The objective of the regularized optimal transport is an $\varepsilon$-strongly convex function, as the function $\mathcal{H}$ is 1-strongly concave : its Hessian is $\partial^2 \mathcal{H}(T) =  -diag(\frac{1}{t_{ij}})$ \text{and} $t_{ij} \le 1.$ Then the regularized problem has a unique optimal solution. Introducing the exponential scaling of the dual variables $u=\exp(\frac{\alpha}{\varepsilon})$ and $v=\exp(\frac{\beta}{\varepsilon})$ as well as the exponential scaling of the cost matrix $K = \exp(\frac{-M_{XY}}{\varepsilon})$, the solution of the regularized optimal transport problem has the form ${T_\varepsilon} = diag(u)Kdiag(v)$. The variables $(u,v)$ must therefore satisfy : $u\odot (Kv) = a  \, \, \, and  \, \, \, v\odot(K^{\mathbf{T}}u) = b$, where $\odot$ corresponds to entrywise multiplication of vectors. This problem is known in the numerical analysis community as the matrix scaling problem, and can be solved efficiently via an iterative procedure : the Sinkhorn-Knopp algorithm \cite{knight2008sinkhorn}, which iteratively update $ u^{(l+1)} = \frac{a}{Kv^{(l)}} , \, \, \, and  \, \, \,    v^{(l+1)} = \frac{b}{K^{\mathbf{T}}u^{(l+1)}},$ initialized with an arbitrary positive vector $v^{(0)}= 1_{m}$. Sinkhorn's algorithm \cite{cuturi2013sinkhorn}, is formally summarized in Algorithm 1 :
\begin{algorithm} \label{sinkhorn}
\caption{Sinkhorn’s algorithm for regularized optimal transport}
\SetKwInOut{Input}{Input}
\SetKwInOut{Parameters}{Parameters}
\SetKwInOut{Output}{Output}
\Parameters{$\varepsilon$}
\Input{$(x_i)_{i=1,...,n}, (y_j)_{j=1,...,m}, a, b$}
$m_{i,j} = \lVert x_i - y_j \rVert^2, \, \, \forall i,j \in \{ 1, ..., n\} \times \{ 1, ..., m\}$ \\
$K = exp(- M_{XY} /\varepsilon)$ \\
Initialize $v  \leftarrow 1_{m}$ \\
\While{not converged}{
 $u \leftarrow \frac{a}{Kv}$  \\
 $v \leftarrow\frac{b}{K^{T}u}$ 
}
\Return {$t_{i,j}=u_iK_{i,j}v_j, \, \, \forall i,j \in \{ 1, ..., n\} \times \{ 1, ..., m\}$}
\end{algorithm}

Note that for a small regularization $\varepsilon$, the unique solution $T_\varepsilon$ of the regularized optimal transport problem converges (with respect to the weak topology) to the optimal solution with maximal entropy within the set of all optimal solutions of $(\mathcal{MK})$ \cite{peyre2019computational}.


\section{Semi-supervised learning}
\label{sec:2}
In traditional machine learning, a categorization has usually been made between two major tasks: supervised and unsupervised learning. In supervised learning settings, algorithms make predictions on some previously unseen data points (test set) using statistical models trained on previously collected labeled data (training set). In unsupervised learning settings, no specific labels are given, instead, one tries to infer some underlying structures (clusters) by relying on some concept of similarity between data points, such that similar samples must be in the same cluster.
\newline
\newline Semi-supervised learning (SSL) \cite{zhu2005semi} is conceptually situated between supervised and unsupervised learning. The goal of semi-supervised learning is to use the large amount of unlabelled instances, as well as a typically smaller set of labelled data points, usually assumed to be sampled from the same or similar distributions, in order to improve the performance that can be obtained either by discarding the unlabeled data and doing classification (supervised learning) or by discarding the labels and doing clustering (unsupervised learning).
\newline
\newline In semi-supervised learning settings, we are presented with a finite ordered set of $l$ labelled data points $\{(x_1,y_1),...,(x_l,y_l)\}$. Each object $(x_i,y_i)$ of this set consists of an data point $x_i$ from a given input space $\mathcal{X}$, and its corresponding label $y_i \in \mathcal{Y} = \{c_1,...,c_K\}$, where $\mathcal{Y}$ is a discrete label set composed by $K$ classes. However, we also have access to a larger collection of $u$ data points $\{x_{l+1},...,x_u\}$, whose labels are unknown. In the remainder, we denote with $X_L$ and $X_U$ respectively the collection of labelled and unlabelled data points, and with $Y_L$ the labels corresponding to $X_L$. 
\newline 
\newline Nevertheless, semi-supervised algorithms work well only under a common assumption: the underlying marginal data distribution $\mathcal{P}(\mathcal{X})$ over the input space $\mathcal{X}$ contains information about the posterior distribution $\mathcal{P}(\mathcal{Y}|\mathcal{X})$ \cite{van2020survey}. When $\mathcal{P}(\mathcal{X})$ contains no information about $\mathcal{P}(\mathcal{Y}|\mathcal{X})$, it is intrinsically inconceivable to improve the accuracy of predictions, in fact, if the marginal data distribution do not influence the posterior distribution, it is inenvisageable to gain information about $\mathcal{P}(\mathcal{Y}|\mathcal{X})$ despite the further knowledge on $\mathcal{P}(\mathcal{X})$ provided by the additional unlabelled data, it might even happen that using the unlabeled data degrades the prediction accuracy by misguiding the inference of $\mathcal{P}(\mathcal{Y}|\mathcal{X})$.  Fortunately, the previously mentioned assumption appears to be satisfied in most of traditional machine learning problems, as is suggested by the successful application of semi-supervised learning methods in numerous real-world  problems.
\newline
\newline Nonetheless, the nature of the causal link between the marginal distribution $\mathcal{P}(\mathcal{X})$ and the posterior distribution $\mathcal{P}(\mathcal{Y}|\mathcal{X})$ and the way of their interaction is not always the same. This has given rise to the semi-supervised learning assumptions, which formalize the types of expected interaction \cite{van2020survey}:
\begin{enumerate}
    \item \textbf{Smoothness assumption :} For two data points $x, x^{'}$ that are close in the input space $\mathcal{X}$, the corresponding labels $y, y^{'}$ should be the same. 
    \item \textbf{Low-density assumption :} The decision boundary should preferably pass through low-density regions in the input space $\mathcal{X}$.
    \item \textbf{Manifold assumption :} The high-dimensional input space $\mathcal{X}$ is constituted of multiple lower-dimensional substructures known as manifolds and samples lying on the same manifold should have the same label.
    \item \textbf{Cluster assumption :} Data points belonging to the same cluster are likely to have the same label.
    \newline
\end{enumerate}
According to the goal of the semi-supervised learning algorithms, we can differentiate between two categories : \textbf{Transductive} and \textbf{Inductive} learning \cite{van2020survey}, which give rise to distinct optimization procedures. The former are  solely  concerned  with obtaining label predictions for the given unlabelled data points, whereas the latter  attempt  to  infer a good classification function that can estimate the label for any instance in the input space, even for previously unseen data points.
\newline
\newline The nature and the objective of transductive methods, make them inherently a perfect illustration of \textbf{Vapnik's principle: when trying to solve some problem, one should not solve a more difficult problem as an intermediate step} \cite{chapelle2009semi}. This principle naturally suggests finding a way to propagate information via direct connections between samples by rising to a graph-based approach. If a graph can be defined in which similar samples are connected, information can then be propagated along its edges. According to Vapnik's principle, this method will allow us to avoid the inference of a classifier on the entire input space and afterwards return the evaluations at the unlabelled points. Graph-based semi-supervised learning methods generally involve three separate steps: graph creation, graph weighting and label propagation. In the first step, vertices (representing instances) in the graph are connected to each other, based on some similarity measure. In the second step, the resulting edges are weighted, yielding an affinity matrix, such that, the stronger the similarity the higher the weight. The first two steps together are commonly referred to as graph construction phase. Once the graph is constructed, it is used in the second phase of label propagation to obtain predictions for the unlabelled data points \cite{subramanya2014graph}. The same already constructed graph will be used in inductive tasks to predict the labels of previously unseen instances.
 
\section{Proposed approach}
\label{sec:3}
\subsection{Optimal Transport Propagation (OTP)}
In this section we show how the transductive  semi-supervised  learning problem can be casted in a principally new way and how can be solved using optimal transport theory.
\newline
\newline
\textbf{Problem setup :} Let $X=\{x_1,...,x_{l+u}\}$ be a set of $l+u$ data points in $\mathbb{R}^d$ and $\mathcal{C} =\{c_1,...,c_K\}$ a discrete label set for $K$ classes. The first $l$ points denoted by $X_L=\{x_1,...,x_l\} \subset \mathbb{R}^d$ are labeled according to $Y_L=\{y_1,...,y_l\}$, where $y_i \in \mathcal{C}$ for every $i \in \{1,...,l\}$, and the remaining data points denoted by $X_U=\{x_{l+1},...,x_{l+u}\} \subset \mathbb{R}^d$ are unlabeled, usually $l \ll u$.
\newline
\newline The goal of transductive semi-supervised learning is to infer the true labels $Y_U$ for the given unlabeled data points using all instances in $X=X_L \cup X_U$ and labels $Y_L$.
\newline
\newline To use optimal transport, we need to express the empirical distributions of $X_L$ and $X_U$ in the formalism of discrete measures respectively as follows :
\begin{equation}
\mu = \sum_{i=1}^l a_{i} \delta_{x_{i}} \, \, \, \text{and} \, \, \, \nu = \sum_{j=l+1}^{l+u} b_{j} \delta_{x_{j}}.
\end{equation}
Emphasized that when the samples $X_L$ and $X_U$ are a collection of independent data points, the weights of all instances in each sample are usually set to be equal :
\begin{equation} a_i = \frac{1}{l}, \, \, \forall i \in \{1,...,l\} \, \, \, \text{and} \, \, \, b_j = \frac{1}{u}, \, \, \forall j \in \{l+1,...,l+u\}.
\end{equation}
\newline \textbf{Proposed approach :} 
Transductive techniques typically use graph-based methods, the common denominator of graph-based methods is the graph construction phase consisting of two steps, the first one lies to model the whole dataset as a graph, where each data point is represented by a vertex, and then forming edges between vertices, and graph weighting step, where a weight is associated with every edge in the graph to provide a measure of similarity between the two vertices joining by each edge. Nonetheless, the most of graph construction methods are either heuristic or very sophisticated, and do not take into account all the geometrical information in the input space. In this paper, we propose a novel natural approach called Optimal Transport Propagation (OTP) to estimate $Y_U$, by constructing a complete bipartite edge-weighted graph  and then propagate labels through its vertices, which can be achieved in two phases:
\newline
\newline \textbf{Phase 1} : Construct a complete bipartite edge-weighted graph  $\mathcal{G}=(\mathcal{V},\mathcal{E},\mathcal{W})$, where $\mathcal{V} = X$ is the vertex set, that can be divided into two disjoint and independent parts $\mathcal{L} = X_L$ and $\mathcal{U} = X_U$, $\mathcal{E} \subset \{\mathcal{L} \times \mathcal{U}\}$ is the edge set, and $\mathcal{W} \in \mathcal{M}_{l,u}(\mathbb{R}^+)$ is the affinity matrix to denote the edges weights, the weight $w_{i,j}$ on edge $e_{i,j} \in \mathcal{E}$ reflects the degree of similarity between $x_i \in X_L$ and $x_j \in X_U$. 
\newline
\newline The first difference we can notice between our label propagation approach and traditional ones is the type of the graph $\mathcal{G}$ and the dimension of the affinity matrix $\mathcal{W}$. Typically, traditional algorithms consider a fully connected graph, that generates a square affinity matrix $\mathcal{W}$, which could be very interesting in the framework of label noise \cite{subramanya2014graph}, i.e. if our goal is to label all the unlabeled instances, and possibly, under the label noise assumption, also to re-label the labeled examples. However, the objective function of these traditional approaches penalize predicted labels that do not match the correct label \cite{van2020survey}, in other words, for labelled data points, the predicted labels should match the true ones, which makes the label noise assumption superfluous. Instead of this paradigm, we adopt a complete bipartite  edge-weighted graph \cite{asratian1998bipartite}, whose vertices can be divided into two disjoint sets, the first set models the labelled data, and the second one, models the unlabelled data. Each edge of the graph has endpoints on differing sets, i.e. there is no edge between two vertices of the same set, and every vertex of the first set is connected to every vertex of the second one. This type of graph, naturally induces a rectangular affinity matrix, containing less elements than the affinity matrix that one could have by considering a fully connected graph : $l \times u < (l+u)
^2$.
\newline
\newline With regard to the graph construction, it must take into consideration that, intuitively, we want data points that are close in the input space $\mathbb{R}^d$ to have similar labels (smoothness assumption), so we need to use some distance that quantitatively defines the closeness between data points. Let $M_{X_LX_U} = (m_{i,j})_{1 \le i\le l,l+1 \le j \le l+u} \in \mathcal{M}_{l,u}(\mathbb{R}^+)$ denotes the matrix of pairwise squared distances between elements of $X_L$ and $X_U$, defined by :
\begin{equation}
m_{i,j} = \lVert x_i - x_j \rVert^2,  \, \, \, \forall i,j  \in \{1, ..., l\} \times \{l+1, ..., l+u \}
\end{equation}
Since our approach satisfy the smoothness assumption, then any two instances that lie close together in the input space $\mathbb{R}^d$ have the same label. Thus, in any densely area of $\mathbb{R}^d$, all instances are expected to have the same label. Therefore, a decision boundary can be constructed that passes only through low-density areas in $\mathbb{R}^d$, thus satisfying the low-density assumption as well.
\newline 
\newline Traditional label propagation approaches \cite{zhu2002learning}\cite{zhou2003learning} create a fully connected weighted graph as mentioned earlier, and a so-called affinity matrix $\mathcal{W}$ is used to denote all the edge weights based on the distance matrix and a Gaussian kernel controlled by a free parameter $\sigma$, in the following way : an edge $e_{i,j}$ is weighted so that the closer the data points $x_i$ and $x_j$ are, the larger the weight $w_{i,j}$ :
\begin{equation}
  w_{i,j} = \exp (-\lVert x_i - x_j \rVert^2 /2 \sigma^2 ), \, \, \, \forall i,j  \in \{1, ..., l+u\}
\end{equation}
Usually, $\sigma$ is determined empirically. However, as mentioned by \cite{zhou2003learning}, it is hard to determine an optimal $\sigma$ if only very few labeled instances are available. Furthermore, even a small perturbation of $\sigma$, as pointed out by \cite{wang2007label}, could make the classification results very different.
\newline
\newline In \cite{wang2007label}, authors suggest to derive another way to construct the affinity matrix $\mathcal{W}$, using the neighborhood information of each point instead of considering pairwise relationships as Eq(10). However, this approach presents two issues : the first one resides in the assumption that all the neighborhoods are linear, in other words each data point can be optimally reconstructed using a linear combination of its neighbors : $x_i = \sum_{j / x_j \in N_{(x_i)}} w_{i,j}x_j$, where $N_{(x_i)}$ represents the neighborhood of $x_i$, and $ w_{i,j}$ is the contribution of $x_j$ to $x_i$, such that $\sum_{j \in N_{(x_i)}} w_{i,j} = 1$ and $w_{i,j} \ge 0$, but it is clear that this linearity assumption is unfortunately not always verified, moreover, the authors mention that it is mainly for computational convenience, the second issue is the choice of the optimal number $k$ of data points constituting the linear neighborhood, typically $k$ is selected by various heuristic techniques, however a variation of $k$ could make the classification results extremely divergent. 
\newline 
\newline To overcome these issues, the new version of the label propagation algorithm proposed in this paper, attempts to exploit all the information in $X$ and the relations between labeled and unlabeled data points in a global vision instead of the pairwise relationships or the local neighborhood information, and without using a Gaussian Kernel. This faculty is ensured naturally by optimal transport, which is a powerful tool for capturing the underlying geometry of the data, by relying on the cost matrix $M_{X_LX_U}$ which encodes the geometry of $X$. Since optimal transport suffers from a computational burden, we can overcome this issue by using the entropy-regularized optimal transport version that can be solved efficiently with Sinkhorn's algorithm proposed in \cite{cuturi2013sinkhorn}. The regularized optimal plan $T^* = (t_{i,j})_{1 \le i\le l,l+1 \le j \le l+u} \in \mathcal{M}_{l,u}(\mathbb{R}^+)$ between the two measures $\mu$ and $\nu$ is the solution of the following problem  : 
\begin{equation}
T^* = \underset{T \in U(a,b)}{argmin} \,\, \langle {T},{M_{X_LX_U}} \rangle _F - \varepsilon \mathcal{H}(T)   ,
\end{equation}
The optimal transport matrix $T^*$ can be seen as a similarity matrix between the two parts $\mathcal{L}$ and $\mathcal{U}$ of the graph $\mathcal{G}$, in fact the elements of $T^*$ provides us with the weights of associations with labeled and unlabelled data points. In other words, the amount of masses $t_{i,j}$ flowing from the mass found at $x_i \in \mathcal{L}$ toward $x_j\in \mathcal{U}$ can be interpreted in our label propagation context as the degree of similarity between $x_i$ and $x_j$ : similar labeled and unlabelled data points correspond to higher value in $T^*$.
\newline
\newline The advantage of our optimal transport similarity matrix is its ability to capture the geometry of the whole $X$ before assigning the weight $t_{i,j}$ to the pair $(x_i,x_j)$, unlike the similarity matrix constructed by the Gaussian kernel that captures information in an bilateral level by considering only the pairwise relationships as \cite{zhu2002learning}\cite{zhou2003learning}, or in an local level by considering the linear neighborhood information as \cite{wang2007label}, these two methods neglect the other interactions which may occur between all the data points in $X$. This geometrical ability allows to the matrix $T^*$ to contain mush more information than the classical similarity matrices. 
\newline
\newline In order to have a class probability interpretation, we column-normalize $T^*$ to get a non-square left-stochastic matrix $P = (p_{i,j})_{1 \le i\le l,l+1 \le j \le l+u} \in \mathcal{M}_{l,u}(\mathbb{R}^+)$, as follows :
\begin{equation}
p_{i,j} = \frac{t_{i,j}}{\sum_i t_{i,j}}, \, \, \, \forall i,j  \in \{1, ..., l\} \times \{l+1, ..., l+u \}. 
\end{equation}
where $p_{i,j},\, \, \,  \forall i,j  \in \{1, ..., l\} \times \{l+1, ..., l+u \}$ is then the probability of jumping from $x_i$ to $x_j$. We consider $P$ as the affinity matrix $\mathcal{W}$. Our intuition is to use this matrix to identify labelled data points who should spread their labels to similar unlabelled data points in the second phase. 
\newline
\newline \textbf{Phase 2}: Propagate labels from the labeled data $X_L$ to the remaining unlabeled data $X_U$ using the bipartite edge-weighted graph constructed in the first phase. We will use an incremental approach to achieve this objective.
\newline
\newline Let $U = (u_{j,k})_{l+1 \le j \le l+u,1 \le k \le K} \in \mathcal{M}_{u,K}(\mathbb{R}^+)$ be a label matrix defined by :
\begin{equation}
u_{j,k} = \sum_{i / x_i \in c_k} p_{i,j} ,\, \, \,  \forall j,k \in \{l+1, ..., l+u \} \times \{1, ..., K\}.
\end{equation}
Note that $U$ is by definition a non-square right-stochastic matrix, and can be understand as a vector-valued function $U : X_U \rightarrow \sum_K$, which assigns a stochastic vector $U_j$ to each unlabeled data point $x_j$. Then for all $j \in \{l+1, ..., l+u\}, \,  x_j$ have soft labels, which can be interpreted as a probability distribution over classes. The choice of the construction of the matrix $U$ in this way, follows the principle : the probability of an unlabeled data point to belong to a class $c_k$ is proportional to its similarity with the representatives of this class. The more the similarity is strong the more the probability of belonging to this class is high.
\newline
\newline The label matrix $U$ will be used to assign a pseudo label for each unlabeled data point in the following way : $ \forall j \in \{l+1, ..., l+u \}$, $x_j$ will take the label corresponding to the class $c_k, k \in \{1, ..., K\}$, which have the largest class-probability $u_{j,k}$ of moving from all the data points belonging to $c_k$ toward $x_j$.
\begin{equation}
\hat{y_{j}} =  \underset{c_{k} \in \mathcal{C}}{argmax} \, u_{j,k}, \, \,  \forall j \in \{l+1, ..., l+u \}.
\end{equation}
Nevertheless, inferring pseudo-labels from matrix $U$ by hard assignment according to Eq(14) in one fell swoop has a displeasing impact : we define pseudo-labels on all unlabeled data points whereas plainly we do not have a steady certainty for each prediction.  This effect, as pointed out by \cite{iscen2019label}, can degrade the performance of the label propagation process. To overcome this issue, we will assign to each pseudo-label $\hat{y_{j}}$ a score $s_j \in  [0, 1]$ reflecting the certainty of the prediction. 
\newline
\newline The certainty score $s_j$ associated with the label prediction $\hat{y_{j}}$ of each data point $x_j \in X_U$ can be constructed as follows : for each unlabelled data point $x_j$, we define a real-valued random variable $Z_j : \mathcal{C} \rightarrow  \mathbb{R}$ defined by $Z_j(c_k) =k$, to associate a numerical value $k$ to the potential label prediction result $c_k$. The probability distribution of the random variable $Z_j$ is encoded in the stochastic vector $U_j$ :
\begin{equation}
\mathbb{P}(Z_j = c_k) =u_{j,k}, \, \, \, \, \, \forall j,k \in \{l+1, ..., l+u \} \times \{1, ..., K\} 
\end{equation}
Then, the certainty score associated with each unlabelled data point can be defined as :
\begin{equation}
s_j = 1 -  \frac{H(Z_j)}{log_2(K)}, \, \, \, \, \, \forall j \in \{l+1, ..., l+u \} 
\end{equation}
where $H$ is Shannon's entropy \cite{shannon2001mathematical}, defined by :
 \begin{equation}
 H(Z_j) = - \sum_k \mathbb{P}(Z_j = c_k)\log_2(\mathbb{P}(Z_j = c_k)) = - \sum_k u_{j,k}\log_2(u_{j,k}). 
 \end{equation}
Since Shannon's entropy $H$ is an uncertainty measure that reach the maximum $log_2(K)$ when all possible events are equiprobable, i.e. $\mathbb{P}(Z_j = c_k) = \frac{1}{K}, \, \, \,  \forall k \in \{1, ..., K \}$ : 
\begin{equation}
\begin{split}
H(Z_j)  & = H(u_{j,1},..., u_{j,K}), \, \, \, \, \, \,  \forall j \in \{l+1, ..., l+u \}  \\
& \le H(\frac{1}{K},..., \frac{1}{K}) \\
& = - \sum_k \frac{1}{K} \log_2(\frac{1}{K})  \\
& =log_2(K) 
\end{split}
\end{equation}
Then, the certainty score $s_j$ is naturally normalized in $[0, 1]$ as mentioned above.
\newline
\newline To control the certainty of the propagation process, we define a confidence threshold $\alpha  \in  [0, 1]$, and for each unlabelled data point $x_j$, if the score $s_j$ is greater than $\alpha$, we assign $x_j$ a pseudo-label $\hat{y_{j}}$ according to Eq(14), and then, we inject $x_j$ into $X_L$, and  $\hat{y_{j}}$ into $Y_L$, otherwise, we maintains $x_j$ in $X_U$. This procedure defines one iteration of our incremental algorithm, in each of its iterations, we modify $X_L$, $Y_L$ and $X_U$, in fact at each iteration, we increase the size of $X_L$, in return, the number of data points in $X_U$ decreases. 
\newline
\newline The effectiveness of a label propagation algorithm depends on the amount of prior information, so enriching $X_L$ at each iteration with new data points, will allow to the instances still in $X_U$ due to the lack of certainty in their label predictions to be labeled, this time with greater certainty.
\newline
\newline We repeat the same whole procedure until convergence, here convergence means that all the data initially in $X_U$ are labelled during the incremental procedure.
\newline
\newline The proposed algorithm, named OTP, is formally summarized in Algorithm 2.
\begin{algorithm}\label{OTP}
\caption{OTP}
\SetKwInOut{Input}{Input}
\SetKwInOut{Parameters}{Parameters}
\SetKwInOut{Output}{Output}
\Parameters{$\varepsilon, \alpha $}
\Input{$X_L, X_U, Y_L$}
\While{not converged}{
Compute the cost matrix $M_{X_LX_U}$ by Eq(9) \\
Solve the regularized Optimal Transport problem in Eq(11) \\
Compute the affinity matrix $P$ by Eq(12)\\
Get the label matrix $U$ by Eq(13) \\
\For{$x_j \in X_U$}{
Compute the certainty score $s_j$ by Eq(16) \\
\eIf{$s_j > \alpha $}{
    Get the pseudo label $\hat{y_{j}}$ by Eq(14)\\
    Inject $x_j$ in $X_L$ \\
    Inject $\hat{y_{j}}$ in $Y_L$ \\
    
}{
Maintain $x_j$ in $X_U$ \\
}
}
}
\Return{$Y_U$}
\end{algorithm}
\begin{figure}[h]
	\centering
\includegraphics[width=1.0\textwidth]{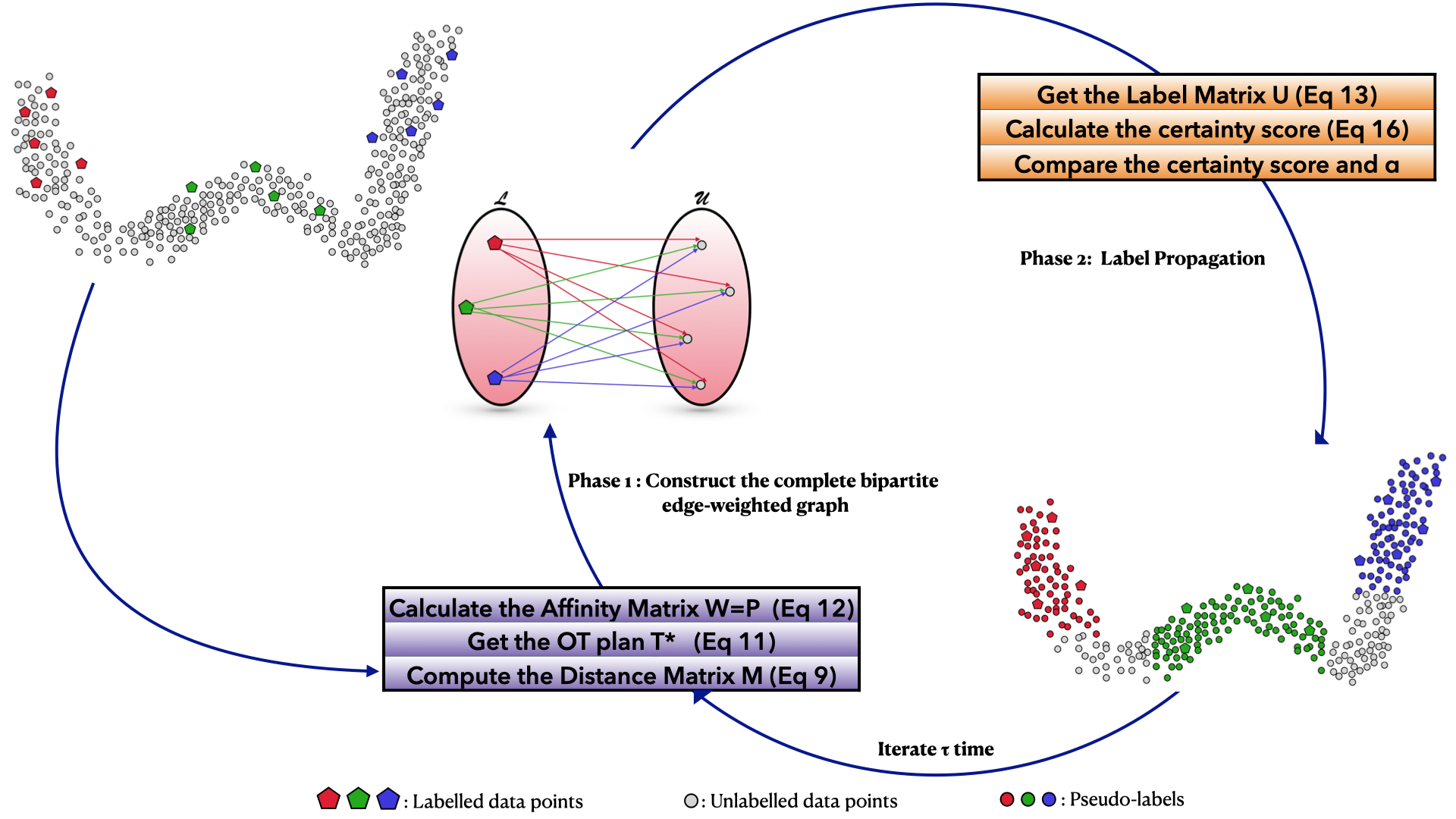}
	\label{fig:pccccs}
	\caption{Overview of the proposed approach. We initiate an incremental process where at each iteration we construct a complete bipartite edge-weighted graph based on the optimal transport plan between the distribution of labelled data points and unlabelled ones, then we propagate labels through the nodes of the graph, from labelled vertices to unlabelled ones, after the evaluation of the certainty of the label predictions.}
\end{figure}
\begin{figure}[h]
	\centering
\includegraphics[width=1.0\textwidth]{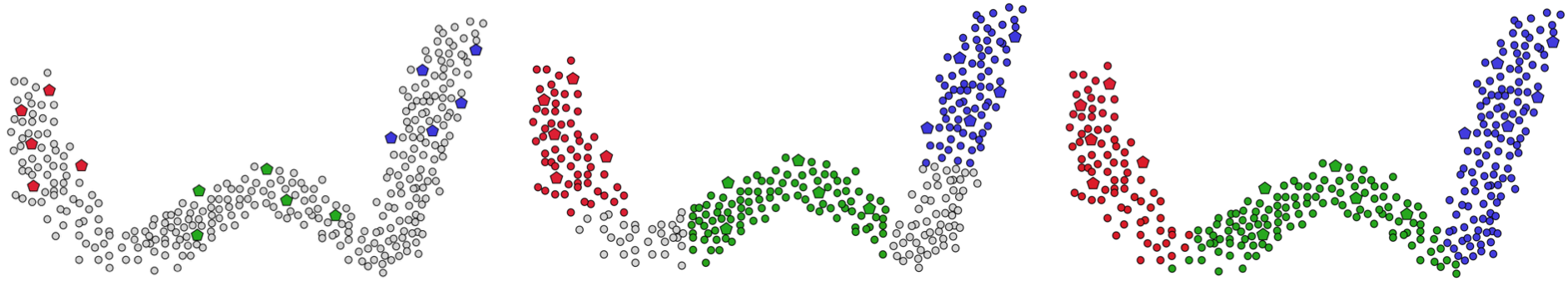}
	\label{fig:pccccccs}
	\caption{The evolution of the label propagation process (from the left to the right): at the initial iteration $t = 0$, at an intermediate iteration $0 < t < \tau$ , and at the last iteration $t = \tau$.
	Pentagon markers correspond to the labeled instances and circles correspond to the unlabeled ones which are gradually pseudo-labeled by OTP. The class is color-coded.}
\end{figure}
\subsection{Convergence analysis of OTP}
As mentioned earlier, the convergence of OTP means that all the data initially in $X_U$ are labelled during the incremental procedure, i.e. when the set $X_L$ absorbs all the instances initially in $X_U$, or in an equivalent way, when $X_U$ is reduced to the empty set $\emptyset$. To analyze the convergence of our approach, we can formulate the evolution of $X_L$ and $X_U$ over time, as follows : 
\newline
\newline Let $m_t$ be the size of the set $X_L$ at time (iteration) $t$, the evolution of $m_t$ is subject to the following nonlinear dynamical system $(R)$ :
\begin{equation} 
(R) \,\,:
\begin{cases} 
m_t = r(m_{t-1}) = m_{t-1} + \zeta_t\\
m_0 = l 
\end{cases}   
\end{equation} 
where $\zeta_t$ is the number of instances in $X_U$ that have been labeled during the iteration $t$. Since in an iteration $t$, we can label all the instances in $X_U$ if the parameter $\alpha$ is too weak, or no point if $\alpha$ is very large, then the terms of the sequence $(\zeta_t)_t$ can vary between $0$ and $u$, and we have $\sum_{t \ge 1} \zeta_t = u$.
\newline
\newline Symmetrically, let $n_t$ be the size of the set $X_U$ at time $t$, this evolution is modeled by the following nonlinear dynamical system $(S)$ :
\begin{equation}
(S) \,\,:
\begin{cases} 
n_t = s(n_{t-1}) = n_{t-1} - \zeta_t\\
n_0 = u 
\end{cases}   
\end{equation}
\newline From a theoretical point of view, our algorithm OTP must converge at the instant $t=\tau$, which verifies :  $m_{\tau} = m_0 + \sum_{t=1}^{\tau} \zeta_t = m_0 + u =l + u$, which corresponds also to $n_{\tau} = n_0 - \sum_{t=1}^{\tau}  \zeta_t = n_0 - u = u -u= 0$.
\newline
\newline OTP will reach the instant $\tau$ in a finite number of iterations, in fact, experiments have shown that a suitable choice of $\alpha$ will allow us to label a large amount $\zeta_t$ of samples in $X_U$ at each iteration $t$, otherwise, it suffices to decrease $\alpha$ in the following way : suppose that at an iteration $t$, we have $h$ unlabeled data points, whose certainty score $s_j$ is lower than the threshold $\alpha$, which means that none of these examples can be labeled according to OTP procedure at the iteration $t$, the solution lies then in decreasing the value of $\alpha$ as follows : 
\begin{equation}
    \alpha \leftarrow \alpha - \min_{x_j \in (X_U)_t}  (\alpha - s_j)
\end{equation}
we denote by $(X_U)_t$ the set of the $h$ points constituting $X_U$ at iteration $t$. Decreasing the value of $\alpha$ in this way, will allow to the point with the greatest certainty score in $(X_U)_t$ to be labeled, and then to migrate from $X_U$ to $X_L$.
\newline
\newline Since integrating this point into $X_L$ can radically change in the next iteration the certainty scores of the $h-1$ data points that are still in $X_U$, we can try to go back to the initial value of $\alpha$, and essay to label normally the $h-1$ instances. Otherwise, if the problem persists, we can apply the same technique of decreasing $\alpha$ to label a new point, and so on until convergence.
\subsection{Induction for Out-of-Sample Data :}  In the previous section we have introduced the main process of OTP, but it is just for the transductive task. In a truly inductive setting, where new examples are given one after the other and a prediction must be given after each example, the use of the transductive algorithm again to get a label prediction for the new instances is very computationally costly, since it needs to be rerun in their entirety, which is unpleasant in many real-world problems, where on-the-fly classification for previously unseen instances is indispensable. 
\newline
\newline In this section, we propose an efficient way to extend OTP for out-of-sample data. In fact, we will fix the transductive predictions $\{y_{l+1},...,y_{l+u}\}$ and based on the objective function of our transductive algorithm OTP we will try to extend the resulting graph to predict the label of previously unseen instances.  
\newline
\newline OTP approach can be casted as the minimization of the cost function $C_{\mathcal{W},l_u}$ in terms of the label function values at the unlabeled data points $x_j \in X_U$ :
\begin{equation}
   C_{\mathcal{W},l_u}(f) = \sum_{x_i \in X_L} \sum_{x_j \in X_U} w_{x_i,x_j} l_u(y_i,f(x_j))
\end{equation}
where $l_u$ is an unsupervised loss function. The cost function in Eq(22) is a smoothness criterion that lies for penalizing differences in the label predictions for connected data points in the graph, which means that a good classifying function should not change too much between similar instances.
\newline
\newline  In order to transform the above transductive algorithm into function induction for out-of-sample data points, we need to use the same type of smoothness criterion as before for a new testing instance $x_{new}$, and then we can optimize the objective function with respect to only the predicted label $\Tilde{f}(x_{new})$ \cite{bengio200611}. The smoothness criterion for a new test point $x_{new}$ becomes then :
\begin{equation}
C^*_{\mathcal{W},l_u}(\Tilde{f}(x_{new}))  =  \sum_{x_i \in X_L \cup X_U} w_{x_i,x_{new}} l_u(y_i,\Tilde{f}(x_{new}))
\end{equation}
If the loss function $l_u$ is convex, e.g. $l_u = (y_i -\Tilde{f}(x_{new}))^2$, then the cost function $C^*_{\mathcal{W},l_u}$ is also convex in $\Tilde{f}(x_{new})$, the label assignment $\Tilde{f}(x_{new})$ minimizing $C^*_{\mathcal{W},l_u}$ is then given by :
\begin{equation}
\Tilde{f}(x_{new}) = \frac{\sum_{x_i \in X_L \cup X_U} w_{x_i,x_{new}} y_i}{\sum_{x_i \in X_L \cup X_U} w_{x_i,x_{new}}}  
\end{equation}
In a binary classification context where $\mathcal{C} =\{+1,-1\}$, the classification problem is transformed into a regression one, in a way that the predicted class of $x_{new}$ is thus $sign(\Tilde{f}(x_{new}))$. 
\newline
\newline It would be very interesting to see what happens when we apply the induction formula (Eq 24) on a point $x_i$ of $X_U$. Ideally, the induction formula must be consistent with the prediction get it by the transduction formula (Eq 22) for an instance $x_i \in X_U$. This is exactly the case in OTP, in fact, induction formula gives the same result as the transduction one over unlabeled points : for $x_{new} = x_k$, $k \in  \{l+1, ..., l+u\} $, we have :
\begin{equation}
\frac{\partial C}{\partial f(x_k)}  = -2 \sum_{x_i \in X_L}  w_{i,k} (y_i - f(x_k))
\end{equation}
$C$ is convex in $f(x_k)$, and is minimized when :
\begin{equation}
\begin{split}
f(x_k) & = \frac{\sum_{x_i \in X_L} w_{x_i,x_k} y_i}{\sum_{x_i \in X_L} w_{x_i,x_k}} \\
& = \frac{\sum_{x_i \in X_L \cup X_U} w_{x_i,x_k} y_i}{\sum_{x_i \in X_L \cup X_U} w_{x_i,x_k}}\, \, \,\, \, \, \, \, \,  \text{since}  \, \, w_{x_i,x_k} =0, \, \,\,\forall x_i \in X_U  \\
& = \tilde{f}(x_k)
\end{split}
\end{equation}
\newline While the most of transductive algorithms have the ability to handle multiple classes, the inductive methods mostly only work in the binary classification setting, where $\mathcal{C} =\{+1,-1\}$. Following the same logic as \cite{delalleau2005efficient}, our optimal transport approach can be adapted and extended accurately for multi-class settings, in the following way : the label $\Tilde{f}(x_{new})$ is given by the weighted majority vote of the others data points in $X = X_L \cup X_U$:
\begin{equation}
\Tilde{f}(x_{new}) = \underset{c_{k} \in \mathcal{C}}{argmax}  \sum_{x_i \in X_L \cup X_U / y_{i} = c_{k}} w_{x_i,x_{new}}
\end{equation}
Our proposed algorithm for the inductive task called Optimal Transport Induction (OTI), is summarized in algorithm 3, where we use the algorithm (OTP) for training and Eq(27) for testing.
\begin{algorithm}\label{OT-ISSL}
\caption{OTI}
\SetKwInOut{Input}{Input}
\SetKwInOut{Parameters}{Parameters}
\SetKwInOut{Output}{Output}
\Parameters{$\varepsilon, \alpha $}
\Input{$x_{new}, X_L, X_U, Y_L$}
\textbf{(1) Training phase}  \\
Get $Y_U$ by (OTP) \\
\textbf{(2) Testing phase}  \\
For a new point $x_{new}$, compute its label $\Tilde{f}(x_{new})$ by Eq(27)  \\
\Return{$\Tilde{f}(x_{new})$}
\end{algorithm}

\section{Experiments}
\label{sec:exp}     
\subsection{Datasets}
The experiment was designed to evaluate the proposed approach on 12 benchmark datasets, which can be downloaded from \footnote{ Datasets are available at \url{https://archive.ics.uci.edu/}}. Details of these  datasets appear in Table 1.
\begin{table}[h]
\caption{Experimental datasets}
\centering
\begin{tabular}{lccc}
\hline
Datasets  \quad \quad   & \#Instances  \quad \quad  & \#Features \quad \quad& \#Classes  \\  \toprule 
Iris                       & 150                     & 4                & 3           \\ 
Wine                       & 178                     & 13                & 3           \\ 
Heart                       & 270                     & 13                & 2         \\ 
Ionosphere                      & 351                     & 34                & 2           \\ 
Dermatology                      & 366                    & 33                & 6          \\
Breast                       & 569                     & 31                & 2           \\ 
WDBC                       & 569                     & 32                & 2
           \\ 
Isolet                       & 1560                     & 617                & 26          \\ 
Waveform                       & 5000                    & 21               & 3  \\ 
Digits                       & 5620                     & 64                & 10           \\ 
Statlog                      & 6435                     & 36               & 6 \\ 
MNIST                       & 10000                     & 784                & 10 \\  \toprule 
\end{tabular}
\label{Data}
\end{table}
\subsection{Evaluation Measures}
Three widely used evaluation measures were employed to evaluate the performance of the proposed approach: the \textbf{Accuracy (ACC)} \cite{liu2019evaluation}, the \textbf{Normalized Mutual Information (NMI)} \cite{dom2012information}, and the \textbf{Adjusted Rand Index (ARI)} \cite{hubert1985comparing}. 
\begin{itemize}
    \item Accuracy (ACC) is the percentage of correctly classified samples, formally, accuracy has the following definition:
\begin{center}
$Accuracy = \frac{\text{Number of correct predictions}}{\text{Total number of predictions}}$
\end{center}
    \item Normalized Mutual Information (NMI) is a normalization of the Mutual Information (MI) score to scale the results between 0 (no mutual information) and 1 (perfect correlation). In this function, mutual information is normalized by some generalized mean of true labels $Y$ and predicted labels $\hat{Y}$.
\begin{center}
$NMI(Y,\hat{Y})= \frac{2I(Y,\hat{Y})}{H(Y)+H(\hat{Y})}$
\end{center}
where $I$ is the mutual information of $Y$ and $\hat{Y}$, defined as:
$I(Y,\hat{Y})= H(Y) - H(Y|\hat{Y})$
with $H$ is the entropy defined by: $H(Y)= \sum_{y}p(y)\log(p(y))$
    \item The rand index is a measure of the similarity between two partitions $A$ and $B$, and is calculated as follows :
\begin{center}
$Rand(A,B) = \frac{a+d}{a+b+c+d}$
\end{center}
where :
$a$ is the number of pairs of elements that are placed in the same cluster in $A$ and in the same cluster in $B$, $b$ denotes the number of pairs of elements in the same cluster in $A$ but not in the same cluster in $B$, $c$ is the number of pairs of elements in the same cluster in $A$ but not in the same cluster in $B$ and $d$ denotes the number of pairs of elements in different clusters in both partitions. The
values $a$ and $d$ can be interpreted as agreements, and $b$ and $c$ as disagreements.
\newline
\newline The Rand index is then “adjusted for chance” into the ARI using the following scheme:
\begin{center}
$ARI = \frac{Rand - ExpectedRand}{maxRand- ExpectedRand}$
\end{center}
The adjusted Rand index is thus ensured to have a value close to 0 for random labeling independently of the number of clusters and samples and exactly 1 when the clustering are identical (up to a permutation).
\end{itemize}
\subsection{Experimental protocol}
The experiment compared the proposed algorithm with three semi-supervised approach, including LP \cite{zhou2003learning} and LS \cite{zhu2002learning}, which are the classical label propagation algorithms, LNP \cite{wang2007label}, which is an improved label propagation algorithm with modified affinity matrix, and the spectral clustering algorithm SC \cite{ng2001spectral} without prior information. 
\newline
\newline To compare these different algorithms, their related parameters were specified as follows : 
\begin{itemize}
    \item The number of clusters $k$ for spectral clustering was set equal to the true number of classes on each dataset. 
    \item Each of the compared algorithms LP, LS and NLP, require a Gaussian kernel controlled by a free parameter $\sigma$ to be specified to construct the affinity matrix, in the comparisons, each of these algorithms was tested
    with different $\sigma$ values, and its best result with the highest ACC, NMI and ARI values on the dataset was selected.
    \item The efficiency of a semi-supervised algorithm depends on the amount of prior information. Therefore, in the experiment, the amount of prior information data was set to 15, 25, and 35 percent of the total number of data points included in a dataset.
    \item The effectiveness of a semi-supervised approach depends also on the quality of prior information. Therefore, in the experiment, given the amount of prior information, all the compared algorithms were run with 10 different sets of prior information to compute the average results for ACC, NMI and ARI on each dataset.
    \item To give an overall vision of the best approach on all the datasets, we define the following score measurement : \begin{center} $\text{SCORE}(A_i) =  \sum_j  \frac{Perf(A_i,D_j)}{max_i Perf(A_i,D_j)}$ \end{center}
    where $Perf$ indicates the performance according to one of the three evaluation measures above of each approach $A_i$ on each data-sets $D_j$. 
\end{itemize}
\subsection{Experimental results}
Tables 2, 3 and 4 list the performance of the different algorithms on all the datasets. These comparisons indicate that the proposed algorithm is superior to the spectral clustering algorithm, this suggests that prior information is able to improve the label propagation effectiveness, this statement is also confirmed by the fact that given the datasets, all the label propagation algorithms show a growth in their performance in parallel with the increase of the amount of prior information. Furthermore, the tables show that the proposed approach is clearly more accurate than LP, LS, and NLP on most tested datasets. However, on some datasets, OTP performed slightly less accurately than LP. The tables also present the proposed score results of each algorithm, which show that the best score belongs to the proposed label propagation approach based on optimal
transport, followed by LS and LP.
\newline
\newline To confirm the superiority of our algorithm over the compared approaches, and especially LP, we suggest to use the Friedman test and Nemenyi test \cite{demvsar2006statistical}. First, algorithms are ranked according to their performance on each dataset, then there are as many rankings as their are datasets. The Friedman test is then conducted to test the null-hypothesis under which all algorithms are equivalent, and in this case their average ranks should be the same. If the null hypothesis is rejected, then the Nemenyi test will be performed. If the average ranks of two approaches differ by at least the critical difference (CD), then it can be concluded that their performances are significantly different. In the Friedman test, we set the significance level $\alpha = 0.05$. Figure 5 shows a critical diagram representing a projection of average ranks of the algorithms on enumerated axis. The algorithms are ordered from left (the best) to the right (the worst) and a thick line connects the groups of algorithms that are not significantly different (for the significance level $\alpha = 5\%$). As shown in figure 5, OTP seem to achieve a big improvement over the other algorithms, in fact, for all evaluation measures, the statistical hypothesis test shows that our approach is more efficient than the compared ones and that the closest method is LS and then LP, which is normal, as both are label propagation approaches, followed by LNP and finally spectral clustering.
\newline
\newline To further highlight the improvement of performances provided by our approach, we are conducting a sensitivity analysis using the Box-Whisker plots \cite{turkey1977exploratory}. Box-Whisker plots are a non-parametric method to represent graphically groups of numerical data through their quartiles, in ordre to study their distributional characteristics. In figure 6, for each evaluation measure, Box-Whisker plots are drawn from the performance of our algorithm and the compared ones over all the tested datasets. To begin with, performances are sorted. Then four equal sized groups are made from the ordered scores. That is, 25\% of all performances are placed in each group. The lines dividing the groups are called quartiles, and the four groups are referred to as quartile groups. Usually we label these groups 1 to 4 starting at the bottom. In a Box-Whisker plot: the ends of the box are the upper and lower quartiles, so the box spans the interquartile range, the median is marked by a vertical line inside the box, the whiskers are the two lines outside the box that extend to the highest and lowest observations. 
\newline
\newline Sensitivity Box-Whisker plots represents a synthesis of the performances into five crucial pieces of information identifiable at a glance: position measurement, dispersion, asymmetry and length of Whiskers. The position measurement is characterized by the dividing line on the median . Dispersion is defined by the length of the Box-Whiskers. Asymmetry is defined as the deviation of the median line from the centre of the Box-Whiskers from the length of the box. The length of the Whiskers is the distance between the ends of the Whiskers in relation to the length of the Box-Whiskers. Outliers are plotted as individual points. 
\newline
\newline Figure 6 confirms the already observed superiority of our algorithm over the others for the three evaluation measures. Indeed, regarding accuracy, we note that the Box-Whisker plot corresponding to OTP is comparatively short, this suggests that, overall, its performance on the different datasets have a high level of agreement with each other, implying a stability comparable to that of LP and LS, and significantly better than that of LNP and SC. For NMI, the Box-Whisker plot corresponding to our approach is much higher than that of LNP and SC, also noting the presence of 2 outliers for LP and LS, these outliers correspond to Heart and Ionosphere datasets, where both approaches have achieved very low scores, on the other hand, there is an absence of outliers for OTP, these indicators confirm the improvement in terms of NMI by our approach. Concerning ARI, we notice that the medians of LP, LS and OTP are all at the same level, however the Box-Whisker plots for this methods show very different distributions of performances, in fact, the Box-Whisker plot of OTP is comparatively short, implying the improvement of the performance of our algorithm in term of ARI  over the other methods and a better stability. 
\newline
\newline All the experimental analysis indicates then, that the performance of the proposed algorithm is higher than the other label propagation algorithms. This result is mainly attributed to the ability of the proposed algorithm to capture mush more information than the previous algorithms thanks to the enhanced affinity matrix constructed by optimal transport. It is equally noteworthy that the effectiveness of the proposed algorithm lies in the fact that the incremental process take advantage of the dependency of semi-supervised algorithms on the amount of prior information, then the enrichment of the labelled set at each iteration with new data points, allows to the unlabelled instances to be labeled with a high certainty, which explains the improvement provided by our approach. Another reason for the superiority of OTP over the other algorithms is its capacity to control the certitude of the label predictions thanks to the certainty score used, which allows instances to be labeled only if we have a high degree of prediction certainty.
\begin{table}[h]
\caption{Accuracy values for semi-supervised methods}
\small
\centering
\begin{tabular}{lcccccc}
\hline
Datasets & Percent  & LP     & LS & LNP   & OTP & SC \\  \toprule 
        & 15\%  & 0.9437 & 0.9453  &0.8852  & \textbf{0.9507} & \\ 
Iris    & 25\%  & 0.9531  & 0.9540 &0.9261  & \textbf{0.9610} & 0.7953\\ 
        & 35\%  & 0.9561  & 0.9571 &0.9392  & \textbf{0.9796}&  \\ 
        
        & 15\%  & \textbf{0.9296}  & 0.9296 & 0.8462& 0.9250 &  \\ 
Wine    & 25\%  & \textbf{0.9417}  & 0.9417 & 0.8597& 0.9343 & 0.8179\\ 
        & 35\%  & \textbf{0.9482}  & 0.9482 & 0.8727& 0.9388&  \\ 
        
        & 15\%  & 0.7261  & 0.7304 &0.5683 & \textbf{0.7696}&  \\ 
Heart  & 25\%   & 0.7734  & 0.7833 &0.6826 & \textbf{0.8424} & 0.3411\\ 
        & 35\%  & 0.8239  & 0.8352 &0.7731 & \textbf{0.8693}& \\

            & 15\%  & 0.8300  & 0.8310 & 0.8051& \textbf{0.8796} &\\ 
Ionosphere   & 25\%  & 0.8439  & 0.8462 & 0.8146& \textbf{0.8871} & 0.4461 \\ 
             & 35\%  & 0.8458  & 0.8476 &0.8293 & \textbf{0.8978} &\\ 
        
              & 15\%  & 0.9324  & 0.9327 & 0.8948& \textbf{0.9488} & \\ 
Dermatology   & 25\%  & 0.9438  & 0.9438 &0.9163 & \textbf{0.9520} & 0.4943\\ 
              & 35\%  & 0.9536  & 0.9536 & 0.9428& \textbf{0.9566}&  \\

        & 15\%  &0.9566  & 0.9566 &0.9153 & \textbf{0.9587} & \\ 
Breast   & 25\%  & 0.9578  & 0.9578 &0.9296 & \textbf{0.9649} & 0.7830 \\ 
         & 35\%  & 0.9649  & 0.9649 &0.9427 & \textbf{0.9730} & \\ 

        & 15\%  & 1.0000  & 1.0000 &0.9568 & \textbf{1.0000}& \\ 
WDBC   & 25\%  & 1.0000  & 1.0000 &0.9879 & \textbf{1.0000} & 0.9682 \\ 
        & 35\%  & 1.0000  & 1.0000 &0.9970 & \textbf{1.0000}& \\

        & 15\%  & 0.7558  & 0.7558 &0.6519 & \textbf{0.7559} &\\ 
Isolet   & 25\%  & \textbf{0.7782}  & 0.7782 &0.6908 & 0.7767 & 0.5385 \\ 
        & 35\%  & \textbf{0.8077}  & 0.8077 &0.7249 & 0.8053 & \\

           & 15\%  & 0.8318  & 0.8334 & 0.7719& \textbf{0.8469}& \\ 
Waveform  & 25\%  & 0.8401  & 0.8419 &0.7892 & \textbf{0.8504} & 0.3842 \\ 
          & 35\%  & 0.8423 & 0.8425 &0.8062 & \textbf{0.8599} &\\

        & 15\%  & 0.9589  & 0.9589 & 0.9363 & \textbf{0.9678}& \\ 
Digits   & 25\%  & 0.9737  & 0.9737 &0.9571 & \textbf{0.9774} & 0.7906 \\ 
        & 35\%  & 0.9801  & 0.9801  &0.9784 & \textbf{0.9827}& \\ 

        & 15\%  & \textbf{0.8740}  & 0.8730 &0.8249 & 0.8516 &\\ 
Statlog   & 25\%  & \textbf{0.8779}  & 0.8771 &0.8371 & 0.8533  & 0.6516\\ 
        & 35\%  & \textbf{0.8831}  & 0.8821  &0.8474 & 0.8538 &\\

        & 15\% & 0.9210 & 0.9218 &0.8247 & \textbf{0.9421} &\\ 
MNIST  & 25\% & 0.9460 & 0.9451 & 0.8371& \textbf{0.9540} &  0.5719\\ 
        & 35\% & 0.9551 & 0.9571 & 0.8408& \textbf{0.9632} &\\ \toprule 
ALL Datasets  & SCORE  & 35.4544  & 35.4975 & 33.3619 & \textbf{35.8855} & 8.1971\\ \toprule 
\end{tabular}
\label{Accuracy}
\end{table}

\begin{table}[h]
\caption{NMI values for semi-supervised methods}
\small
\centering
\begin{tabular}{lcccccc}
\hline
Datasets & Percent  & LP     & LS & LNP  & OTP  & SC \\ \toprule 
        & 15\%  & 0.8412  & 0.8442  & 0.7534 & \textbf{0.8447} \\ 
Iris    & 25\%  & 0.8584  & 0.8621  & 0.8269& \textbf{0.8667}  & 0.7980\\ 
        & 35\%  & 0.8621  & 0.8649 & 0.8314 & \textbf{0.8852} \\ 
        
        & 15\%  & \textbf{0.7821}  & 0.7821 & 0.6815 & 0.7384 \\ 
Wine    & 25\%  & \textbf{0.8127}  & 0.8127 & 0.7573 & 0.7790 & 0.7808 \\ 
        & 35\%  & \textbf{0.8289}  & 0.8289 & 0.7897 & 0.7963 \\ 
        
        & 15\%  & 0.1519  & 0.1575 &0.1091& \textbf{0.2181} \\ 
Heart   & 25\%   & 0.2291  & 0.2472 &0.1432& \textbf{0.3683} & 0.1880 \\ 
        & 35\%  & 0.3313  & 0.3546 &0.2718& \textbf{0.4374} \\ 

             & 15\%  & 0.3502  & 0.3535 &0.3256& \textbf{0.4676} \\ 
Ionosphere   & 25\%  & 0.3848  & 0.3911 &0.3572& \textbf{0.5000} &  0.2938\\ 
             & 35\%  & 0.3972  & 0.4014 &0.3725& \textbf{0.5383}\\ 
        
              & 15\%  & 0.8770  & 0.8779 &0.8349& \textbf{0.8935} \\ 
Dermatology   & 25\%  & 0.8932  & 0.8932 &0.8692& \textbf{0.9033} & 0.6665 \\ 
              & 35\%  & 0.9128  & 0.9128 &0.8959& \textbf{0.9164} \\

        & 15\%  & 0.7340  & 0.7360 &0.6971& \textbf{0.7449} \\ 
Breast   & 25\%  & 0.7451  & 0.7465 &0.7192& \textbf{0.7550} & 0.6418 \\ 
         & 35\%  & 0.7909  & 0.7909 &0.7706& \textbf{0.8106} \\

        & 15\%  & 1.0000  & 1.0000 &0.9049& \textbf{1.0000} \\ 
WDBC   & 25\%  & 1.0000  & 1.0000 &0.9347& \textbf{1.0000} & 0.9163\\ 
        & 35\%  & 1.0000  & 1.0000 &0.9715& \textbf{1.0000} \\ 
        
        & 15\%  &\textbf{ 0.7785}  & 0.7785 &0.7184& 0.7657 \\ 
Isolet   & 25\%  & \textbf{0.7987}  & 0.7987 &0.7503& 0.7852 & 0.7545\\ 
        & 35\%  & \textbf{0.8210}  & 0.8210 &0.7869& 0.8077 \\  
        
          & 15\%  & 0.4950  & 0.5009 &0.4628& \textbf{0.5256} \\ 
Waveform  & 25\%  & 0.5124  & 0.5192 &0.4763 &\textbf{0.5319} & 0.3646 \\ 
          & 35\%  & 0.5192  & 0.5229 &0.4807& \textbf{0.5421} \\

        & 15\%  & 0.9150  & 0.9150 &0.8891& \textbf{0.9290} \\ 
Digits   & 25\%  & 0.9443  & 0.9443 &0.9268& \textbf{0.9489} & 0.8483 \\ 
        & 35\%  & 0.9570  & 0.9570 &0.9318& \textbf{0.9607} \\ 

        & 15\%  & \textbf{0.7396}  & 0.7383 &0.6792& 0.6753 \\ 
Statlog  & 25\%  & \textbf{0.7483}  & 0.7477 &0.6859 &0.6800 & 0.6139 \\ 
        & 35\%  & \textbf{0.7572}  & 0.7571 &0.6907& 0.6821 \\ 
        
        & 15\% & 0.8019 & 0.8028 & 0.7759& \textbf{0.8177} \\
MNIST  & 25\% & 0.8389 & 0.8367 &0.7931 &\textbf{0.8442} & 0.6321  \\
        & 35\% & 0.8542 & 0.8599 &0.8136& \textbf{0.8730} \\ \toprule 
ALL Datasets  & SCORE  & 33.9980  & 34.2042 & 31.6326 & \textbf{35.5237} & 9.5770 \\ \toprule 
\end{tabular}
\label{NMI}
\end{table}
\begin{table}[h]
\caption{ARI values for semi-supervised methods}
\small
\centering
\begin{tabular}{lcccccc}
\hline
Datasets & Percent  & LP     & LS   & LNP &  OTP & SC \\ \toprule 
        & 15\%  & 0.8453  & 0.8492 &0.7861 & \textbf{0.8621 }\\ 
Iris    & 25\%  & 0.8680  & 0.8704 &0.8321& \textbf{0.8884} & 0.7455 \\ 
        & 35\%  & 0.8754  & 0.8783 &0.8424& \textbf{0.9027} \\ 
        
        & 15\%  & \textbf{0.7936} & 0.7936 &0.7148& 0.7814 \\ 
Wine    & 25\%  & \textbf{0.8267}& 0.8267 &0.7346& 0.8050 &  0.7912\\ 
        & 35\%  & \textbf{0.8455} & 0.8455 &0.7741 & 0.8192\\ 
        
        & 15\%  & 0.2110  & 0.2190 &0.1562& \textbf{0.2875} \\ 
Heart  & 25\%   & 0.3176  & 0.2955 &0.2283& \textbf{0.4662} & 0.2031 \\ 
        & 35\%  & 0.4163  & 0.4464 &0.3688& \textbf{0.5430} \\ 
        
             & 15\%  & 0.4221  & 0.4248& 0.3998 & \textbf{0.5723} \\ 
Ionosphere   & 25\%  & 0.4606  & 0.4673& 0.4324& \textbf{0.5927} & 0.3971\\ 
             & 35\%  & 0.4650  & 0.4702&0.4418 & \textbf{0.6281} \\

               & 15\%  & 0.8807  & 0.8813 & 0.8438& \textbf{0.8996} \\ 
Dermatology   & 25\%  & 0.8972  & 0.8972 &0.8751& \textbf{0.9093} & 0.4783 \\ 
              & 35\%  & 0.9146  & 0.9146 &0.9007& \textbf{0.9218} \\

        & 15\%  & 0.8328  & 0.8327 &0.7956& \textbf{0.8404} \\ 
Breast   & 25\%  & 0.8371  & 0.8284 &0.8039& \textbf{0.8636} & 0.7018 \\ 
         & 35\%  & 0.8632  & 0.8632 &0.8413& \textbf{0.8940}\\

        & 15\%  & 1.0000  & 1.0000 &0.9349& \textbf{1.0000} \\ 
WDBC   & 25\%  & 1.0000  & 1.0000 &0.9691 &\textbf{1.0000}  & 0.9565 \\ 
        & 35\%  & 1.0000  & 1.0000 &0.9905& \textbf{1.0000} \\  
        
        & 15\%  & \textbf{0.6002}  & 0.6002 & 0.5064 & 0.5998 \\ 
Isolet   & 25\%  & \textbf{0.6333} & 0.6332 &0.5526 & 0.6299 & 0.5284 \\ 
        & 35\%  & \textbf{0.6735}  & 0.6735 &0.5992& 0.6683 \\

          & 15\%  & 0.5639  & 0.5678 &0.5163& \textbf{0.5945} \\ 
Waveform  & 25\%  & 0.5819  & 0.5864 &0.5279& \textbf{0.6031} & 0.3788 \\ 
          & 35\%  & 0.5870  & 0.5880 &0.5342& \textbf{0.6182} \\

        & 15\%  & 0.9126  & 0.9127 &0.8993& \textbf{0.9306} \\ 
Digits   & 25\%  & 0.9432  & 0.9432 &0.9287& \textbf{0.9508} & 0.7846 \\ 
        & 35\%  & 0.9567  & 0.9567 &0.9407& \textbf{0.9621} \\ 
        
        & 15\%  & \textbf{0.7658}  & 0.7640 &0.7167 &0.7122 \\ 
Statlog  & 25\%  & \textbf{0.7730}  & 0.7714 &0.7318& 0.7284 & 0.6031 \\ 
        & 35\%  & \textbf{0.7820}  & 0.7806 &0.7391& 0.7336 \\ 

        & 15\% & 0.7930 & 0.7944 &0.7697& \textbf{0.8393} \\
MNIST  & 25\% &  0.8487 & 0.8466 &0.8152& \textbf{0.8685} & 0.5153 \\
& 35\% & 0.8721 & 0.8777 &0.8438& \textbf{0.8935} \\ \toprule 
ALL Datasets  & SCORE  & 33.9814  & 34.0574 & 31.6888 & \textbf{35.7239} & 8.8581\\ \toprule 
\end{tabular}
\label{ARI}
\end{table} 
\begin{figure}[h]
\centering
{\includegraphics[clip,width=0.65\columnwidth]{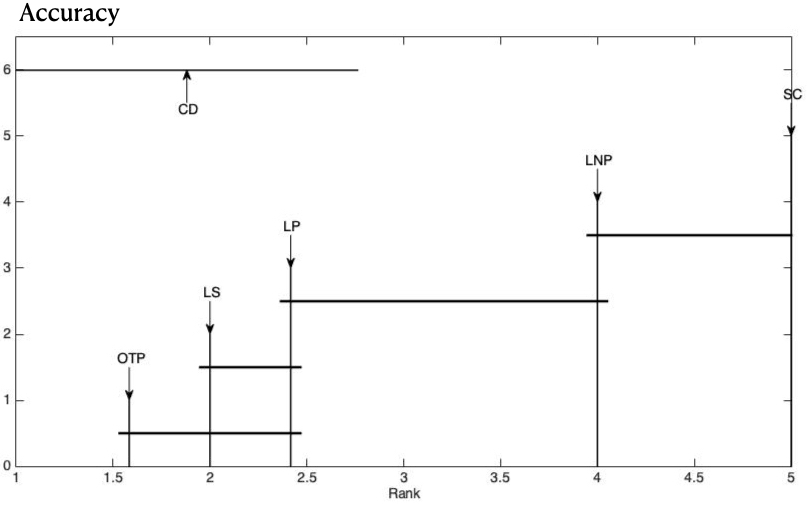}}
{\includegraphics[clip,width=0.65\columnwidth]{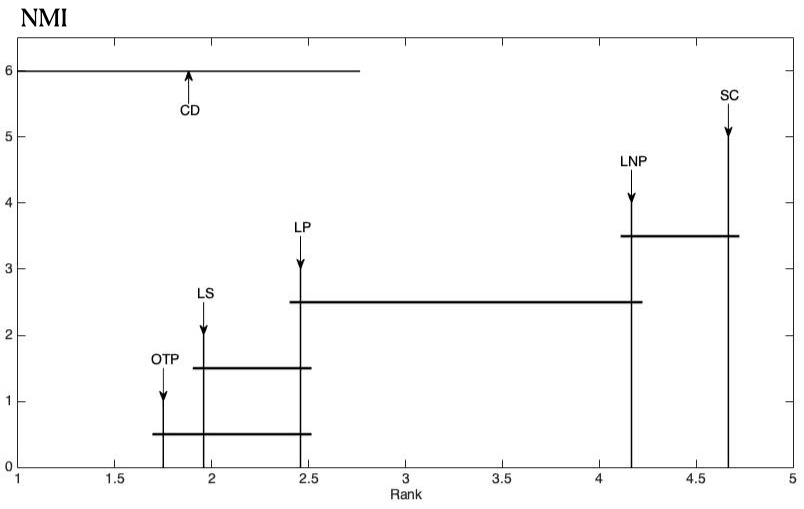}}
{\includegraphics[clip,width=0.65\columnwidth]{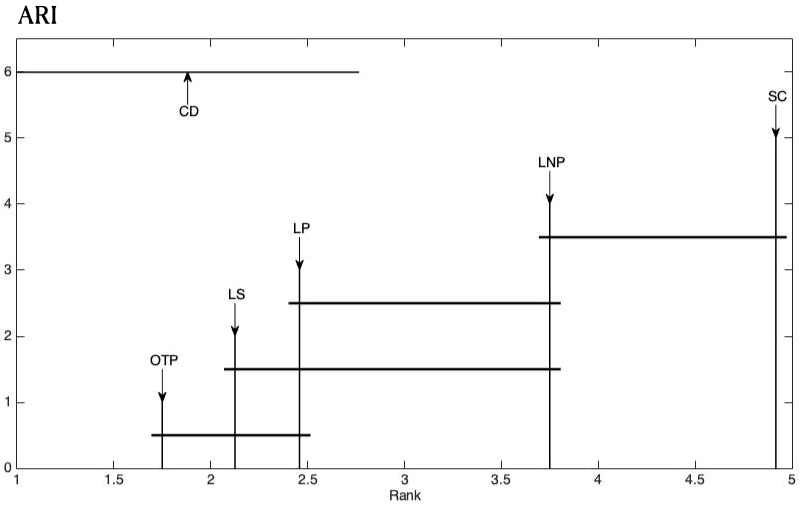}}
\caption{Friedman and Nemenyi test for comparing multiple approaches over multiple datasets using multiple evaluation measures : approaches are ordered from left (the best) to right (the worst)}
\end{figure}
\begin{figure}[h]
\centering
{\includegraphics[clip,width=0.5\columnwidth]{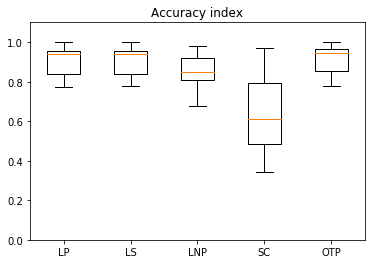}}
{\includegraphics[clip,width=0.5\columnwidth]{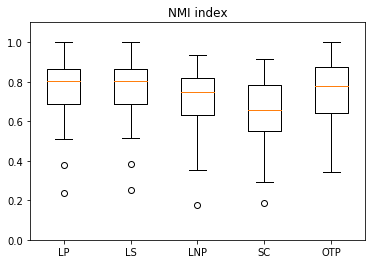}}
{\includegraphics[clip,width=0.5\columnwidth]{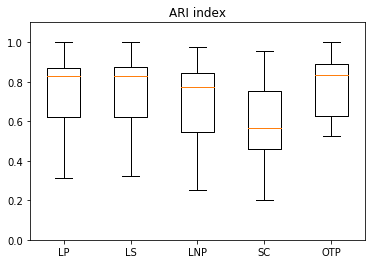}}
\caption{Sensitivity analysis of the multiple approaches using Box-Whiskers plots}
\end{figure}
\restoregeometry
\section{Conclusion and future works}
In this paper, we propose a new label propagation algorithm, named OTP. In the proposed algorithm, the optimal transport plan between the empirical measures of labelled and unlabelled instances is used to construct an enhanced affinity matrix to capture the entire geometry of the underlying space. An incremental process is used to propagate labels from labelled data to unlabelled ones, the process is guided by a certainty score to assure the certitude of predictions. The proposed algorithm not only solves the transductive task, it is also able to be extended efficiently to make predictions for out-of-sample data OTI. Finally, extensive experiments were conducted to demonstrate the effectiveness of the proposed algorithm compared to other semi-supervised learning algorithms. The experimental results indicated that the effectiveness of the proposed algorithm is superior to that of the other algorithms on most of the tested datasets.
\newline
\newline This study mainly focused on label propagation and the generalization to out-of-sample data. In future, we aim to use the predicted labels inferred by our algorithm in conjunction with the initial labelled data to train a convolutional neural network model in a semi-supervised learning fashion. Furthermore, we plan to develop a theoretical analysis of semi-supervised learning with optimal transport theory.

\bibliographystyle{plain}
\bibliography{Mourad.bib}

\end{document}